\documentclass[conference]{IEEEtran}
\IEEEoverridecommandlockouts
\usepackage{cite}
\usepackage{amsmath,amssymb,amsfonts}
\usepackage{algorithmic}
\usepackage{graphicx}
\usepackage{textcomp}
\usepackage{xcolor}
\def\BibTeX{{\rm B\kern-.05em{\sc i\kern-.025em b}\kern-.08em
    T\kern-.1667em\lower.7ex\hbox{E}\kern-.125emX}}
\begin{document}

\title{Exploratory Data Analysis on Code-mixed Misogynistic Comments}

\author{\IEEEauthorblockN{1\textsuperscript{st} Sargam Yadav}
\IEEEauthorblockA{\textit{School of Informatics and Creative Arts} \\
\textit{Dundalk Institute of Technology}\\
Dundalk, Ireland \\
sargam.yadav@dkit.ie}
\and
\IEEEauthorblockN{2\textsuperscript{nd} Abhishek Kaushik}
\IEEEauthorblockA{\textit{School of Informatics and Creative Arts} \\
\textit{Dundalk Institute of Technology}\\
Dundalk, Ireland \\
Abhishek.Kaushik@dkit.ie}
\and
\IEEEauthorblockN{3\textsuperscript{rd} Kevin Mc Daid}
\IEEEauthorblockA{\textit{School of Informatics and Creative Arts} \\
\textit{Dundalk Institute of Technology}\\
Dundalk, Ireland \\
kevin.mcdaid@dkit.ie}

}

\maketitle

\begin{abstract}
The problems of online hate speech and cyberbullying have significantly worsened since the increase in popularity of social media platforms such as YouTube and Twitter (X). Natural Language Processing (NLP) techniques have proven to provide a great advantage in automatic filtering such toxic content. Women are disproportionately more likely to be victims of online abuse. However, there appears to be a lack of studies that tackle misogyny detection in under-resourced languages. In this short paper \footnote{This paper is accepted in the 16th ISDSI-Global Conference 2023 https://isdsi2023.iimranchi.ac.in/}, we present a novel dataset of YouTube comments in mix-code Hinglish collected from YouTube videos which have been weak labelled as `Misogynistic' and `Non-misogynistic'. Pre-processing and Exploratory Data Analysis (EDA) techniques have been applied on the dataset to gain insights on its characteristics. The process has provided a better understanding of the dataset through sentiment scores, word clouds, etc.   
\end{abstract}


\section{Introduction}

In the era of the internet and digital communication, social media platforms have been integrated as a viable component in many companies, businesses, and even the daily lives of people. They serve as the hub of connectivity and communication. However, this increase in interconnectedness has brought about several new issues such as cyberbullying, online hate speech \cite{bohra2018dataset}, and misinformation \cite{moreno2019applying}. Platforms employ techniques such as keyword filtering and manual content moderation to remove hateful and offensive content. However, this approach generally proves to be insufficient, especially for under-resourced and mix-code languages \cite{mandl2019overview}. This is because while detecting hate speech, it is important to keep the context of the conversation in mind \cite{frenda2019online}. Users from multicultural and multilingual countries generally combine their local languages with English in online posts. For example, `Hinglish' is a combination of Hindi and English, and consists of Romanized Hindi text mixed with English words \cite{mathur2018detecting}.

%Artificial intelligence models such as BERT have recently shown outstanding performance on several NLP tasks, including hate speech detection. Recent studies have used models such Transformers \cite{kirk2023semeval}, for hate speech detection, and the results were promising for different languages \cite{mandl2021overview}. %Women are generally targeted more for online harassment, with the consequences being anxiety, depression, withdrawal from social media and public discourse, etc. The number of studies that detect misogyny in online comments and posts is very low, with even fewer studies tackling the problem in mix-code languages due to unavailability of robust datasets, language-specific tools, etc. 

In this paper, we present a novel dataset for misogyny detection in code-switched Hinglish. The dataset was collected from YouTube videos and has been weakly annotated by a single annotator into the following two categories: Misogynistic (`MGY') and Not-misogynistic (NOT). %The misogynistic comments are further classified into the following 9 labels: 'Derailing', 'Sexual Harassment \& Threats of Violence', 'Stereotyping \& Objectification', 'Minimization \& Trivialization', 'Religion-based', 'What aboutism', 'Shaming', 'Moral Policing', 'Victim Blaming'. 
EDA has been performed on the dataset to understand its nature and characteristics, find patterns, detect anomalies, gain insight, and receive feedback. Modelling decisions will be made based on the findings of the study. 
The rest of the paper is structured as follows: Section \ref{mot} discusses the motivation behind the conducting the study, the hypothesis, and research questions. Section \ref{lit} discusses the relevant literature in the field of misogyny and hate speech detection in code-mixed and other languages. Section \ref{dataset} provides details of the dataset. In Section \ref{eda}, EDA is performed on the dataset to find patterns in the data. In Section \ref{disc}, we discuss the findings and insights from the study and how they answer the research questions. Section \ref{conc} concludes the study.

\section{Motivation} \label{mot}
The motivation behind conducting this study is to provide a new dataset of 2229 YouTube comments for misogyny detection in Hinglish, perform exploratory data analysis on the dataset, and gain insight from the findings to guide future modelling decisions. There is a lack of robust datasets for misogyny detection in mix-code languages \cite{satapara2021overview}. Data was gathered with established ethical approval. The dataset has only one annotator, and another one is currently working on annotation, hence it is only sparsely annotated. The main reason for submitting the paper was to solicit expert feedback on the study's workflow. %Artificial intelligence models can be greatly beneficial towards filtering hate speech and protecting targets from continued harassment \cite{fersini2018overview}. 

\textit{Hypothesis:} EDA techniques can derive valuable insights from a code-mixed Hinglish dataset for misogyny detection and find patterns and relationships in the comments and labels.

\begin{enumerate}
    \item Research Question 1: Can EDA techniques help uncover valuable insights from a dataset of Hinglish comments for the misogyny detection?
    \item Research Question 2: What valuable insights and patterns are uncovered? Which EDA techniques provide valuable insights?
\end{enumerate}

\section{Literature Review} \label{lit}
In this section, we will discuss the relevant literature in the fields of misogyny detection and code-mixed languages.

\subsection{Misogyny Detection}
There has been some effort in misogyny detection in languages other than English. Fersini et al. \cite{fersini2018overview} conducted the Automatic misogyny identification (AMI) in Italian and English. %The shared task was divided into two subtasks: Subtask A (Misogyny identification) and Subtask B (type of misogyny). The labels for the subtask B were as follows: Stereotype \& Objectification, Dominance, Derailing, Sexual Harassment \& Threats of Violence, Discredit. The taxonomy in the current study has been inspired by this approach. Similarly, the AMI shared tasks at IberEval 2018 \cite{fersini2018ibereval} and the AMI task at Evalita (2020) \cite{fersini2020ami} followed the same taxonomy and achieved the highest scores using approaches such as BERT-based models, deep learning models, etc. 
Anzovino et al. \cite{anzovino2018automatic} used a similar taxonomy for misogyny detection and achieved the highest score using a Linear Support Vector Machine. Kirk et al. \cite{kirk2023semeval} conducted the Explainable Detection of Online Sexism at SemEval-2023 that consisted of 3 subtasks. Subtask 3 consisted of 11 fine-grained vectors. The highest scores were achieved by Transformer based models such as DeBERTa-v3. %Basile et al. \cite{basile2019semeval} presented the HatEval task of SemEval 2019 which focused on detection of hate comments towards women and immigrants. Frenda et al. \cite{frenda2019online} performed a study to analysis misogynistic and sexist comments, and discovered syntactic similarities between the two. 

\subsection{Code-mixed Languages}
The efforts in hate speech detection for languages such as Marathi, Hindi, and more, have seen a dramatic rise \cite{mandl2019overview} \cite{mandl2020overview}. %The HASOC (Hate speech and Offensive Content Identification) subtasks \cite{mandl2019overview} \cite{mandl2020overview} \cite{mandl2021overview} are conducted yearly to detect hate speech and offensive content from datasets comments of English, German, and other under-sourced languages. 
The Identification of Conversational Hate-Speech in Code-Mixed Languages (ICHCL) shared task focused mainly on Hindi and Hinglish. Similarly, the TRAC subtask \cite{kumar2018benchmarking} was also conducted in mix-code Hinglish. Other works that conduct sentiment analysis on mix-code data include the study done by Kaur et al. \cite{kaur2019cooking}, where they implemented machine learning models and different feature engineering techniques on a YouTube cookery dataset in mix-code Hinglish. Donthula et al. \cite{donthula2019man}
continued the experimentation by using deep learning models, contextual embedding techniques, and statistical methods such as TF-IDF and CountVec. In the study done by Yadav et al. \cite{yadav2021cooking}, BERT-based models were used to perform sentiment analysis on the cookery dataset. This approach was expanded on by Yadav et al. \cite{yadav2022contextualized}, who utilised contextual embeddings from BERT-based models such as RoBERTa. This approach outperformed the approach taken by Kaur et al. \cite{kaur2019cooking} and Yadav et al. \cite{yadav2021cooking}.

\section{Dataset} \label{dataset}
In this section, we will discuss the details of the dataset. The dataset was collected from YouTube videos discussing current social issues in India. Table \ref{tab:video} provides a list of the videos, the channel names, and other information regarding the videos. All comments that are not in English or Romanized Hindi have been removed from the dataset. Any replies to comment that did not form a logical sentence in itself have also been removed. Table \ref{tab:label_counts} depicts the number of comments belonging to each of the classes. The classes are highly imbalanced, as is usually the case in hate speech and misogyny detection \cite{fersini2018overview}, since there are very few positive samples in the real world. The `MGY' class consists of only 181 comments. 

\begin{table*}[t]
    \centering
    \begin{tabular}{p{4cm}p{1.5cm}p{2cm}p{2cm}p{2cm}p{2.5cm}p{2cm}}
    \hline
       \textbf{ Name} & \textbf{Channel} & \textbf{No. of Subscribers} & \textbf{Date Uploaded} & \textbf{No. of views} & \textbf{Extracted comments} & \textbf{Final comments}\\
        \hline
       Know why Delhi is called the Rape Capital & ThePrint & 2.32 million &Dec 15, 2019&987,073 & 11,344 & 1149\\
        
         Use vs Abuse - Debate over declaring Marital Rape a crime & The Deshbhakt & 3.43 million & Jan 26, 2022 & 316,794 &4362& 621\\
       
         What people think about bringing the Marital rape law in India? & InUth & 44.8K& Jul 4, 2022 & 26,724 & 384 & 92 \\
         
         Misogyny in Indian Politics: Time to end this disease! & The Deshbhakt & 3.43
         million & Jan 19, 2022 & 441,144 & 1702 & 362\\
         \hline
    \end{tabular}
    \caption{Video Titles}
    \label{tab:video}
\end{table*}

\begin{table}
    \centering
    \begin{tabular}{c|c}
    \hline
        \textbf{Label name} & \textbf{Count}  \\
        \hline
        NOT &2048  \\
        MGY & 181 \\
         \hline
         Total & 2,229 \\
         \hline
    \end{tabular}
    \caption{Label counts}
    \label{tab:label_counts}
\end{table}

%\begin{table}[]
 %  \centering
   % \begin{tabular}{|c|c|}
  %  \hline
   %      Label Name & Count  \\
    %     \hline
     %    Derailing & 153 \\
      %   Sexual Harassment and Threats of Violence & 4 \\
      %   Stereotyping and Objectification & 125\\
      %   Minimization and Trivialization & 157 \\
      %   Religion-based & 4\\
      %   What aboutism & 23 \\
       %  Shaming & 30\\
       %  Moral Policing & 36\\
        % Victim Blaming & 50\\
        % \hline
        % Total & 184\\
        % \hline
   % \end{tabular}
   % \caption{Labels}
   % \label{tab:labels}
%\end{table}

%Table \ref{tab:labels} displays the total number of samples in each of the nine labels of the class `MGY'. Each comment may have one or more labels. Label 'Minimization and Trivialization' has the highest number of samples while `Religion-based' has the lowest.

\section{Exploratory Data Analysis} \label{eda}
In this section, we will discuss the details of the research methodology and analyse the results.
\subsection{Pre-processing}

Data cleaning and pre-processing was performed on the dataset before using EDA techniques. All non-ASCII characters, punctuation, stopwords, non-English characters, hyperlinks, extra spaces and lines, were removed from the dataset. Stemming was performed using Snowball Stemmer.

\subsection{Analysis}
We will now discuss the various EDA techniques and their results on the dataset. The most common tokens in the dataset are: `i' (888), `rape' (682), `women' (661), `men' (499), and `hai' (492). The average number of characters in each comment are 115.22 and the average number of words are 20.96.

\begin{figure}
\includegraphics[width = 1\linewidth]{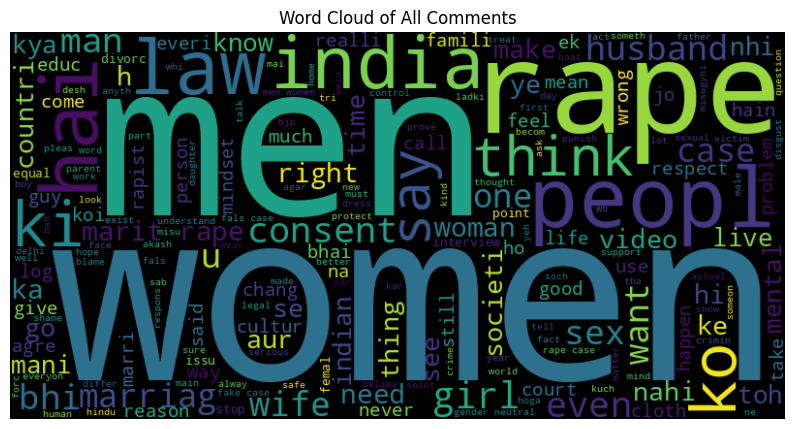}
    \caption{Word Cloud for All Comments}
    \label{fig:wordcloud}
\end{figure}

Fig \ref{fig:wordcloud} displays a word cloud for all the comments in the dataset. The most frequent words in the dataset are `women', `men', `law', `rape', etc. Table \ref{tab:word_Count} shows the average word count for the sequences in each class. Comments belonging to the `MGY' class are 1.6 times longer on an average than comments belonging to the `NOT' class. %Figure \ref{fig:noncloud} displays a word cloud for the comments that belong to the 'NOT' class. Similar to the word cloud for class 'MGY', some of the important words Therefore, there is not much distinction to be made between the two classes based on the presence of unique keywords. 

\begin{table}
    \centering
    \begin{tabular}{c|c|c}
    \hline
     &  \textbf{MGY} & \textbf{NOT} \\
     \hline
      \textbf{Average Word Count}   &  32.72 & 19.91\\
      \hline
    \end{tabular}
    \caption{Average Word Count for Each Class}
    \label{tab:word_Count}
\end{table}

%\begin{figure}[!ht] \includegraphics[width = 1\linewidth]{figures/non.png}
 %   \caption{non word cloud}
  % \label{fig:noncloud}
%\end{figure}

%\begin{figure}[htbp]

%\includegraphics[width = 1.1\linewidth]{figures/labelcorre.png}
 %   \caption{Label Correlation Heatmap}
  %  \label{fig:heatmap}
%\end{figure}
%Figure \ref{fig:heatmap} displays the correlation between the different labels of the 'MGY' category. The highest positive correlation is 0.58 between the labels 'Moral Policing' and 'Victim Blaming', implying that both these labels occur together the most frequently. The lowest correlation is -0.25 between the label 'Sexual Harassment \& Threats of Violence' and the labels 'Derailing' and 'Minimization \& and Trivialization'. There are very few samples (4) for the 'Sexual Harassment and Threats of Violence', which can be the cause of negative correlation.

\begin{figure}
\includegraphics[width = 1\linewidth]{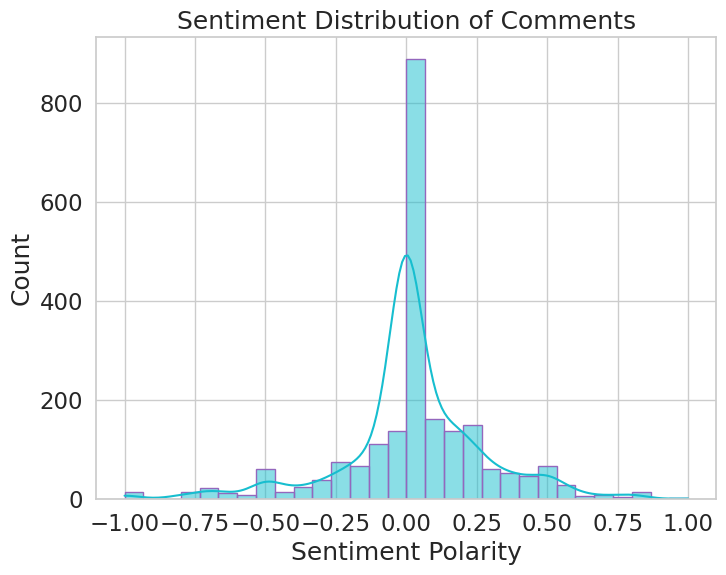}
    \caption{Histogram of Sentiment Polarity vs the Comment Count}
    \label{fig:polarity}
\end{figure}

\begin{figure}
\includegraphics[width = 1\linewidth]{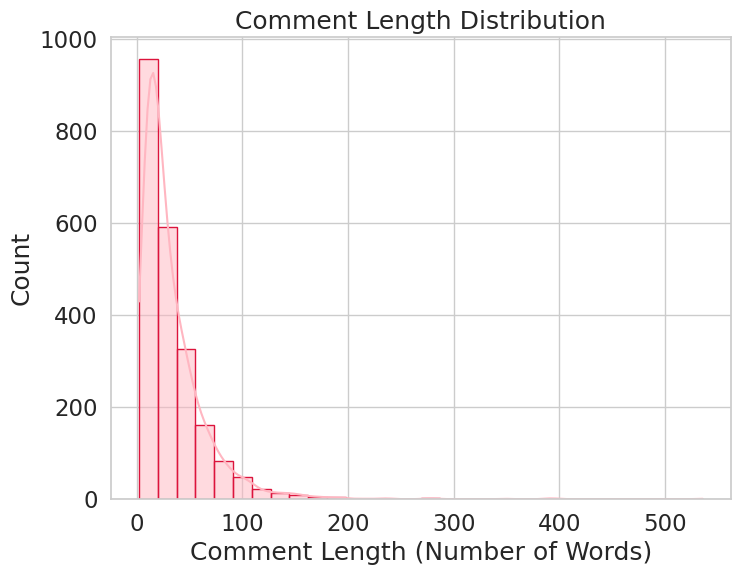}
    \caption{Histogram of Comment Length vs the Comment Count}
    \label{fig:commentlength}
\end{figure}

Fig \ref{fig:polarity} displays a histogram that shows the distribution of sentiment polarity score for all the comments in the dataset. The x-axis represents the sentiment analysis scores of the comments obtained using TextBlob, and ranges from -1 (lowest) to 1 (highest). The y-axis displays the total count of comments. Each of the bins depicts a range of polarity scores. The histogram is slightly skewed to the right, depicting that the number of comments with a positive score is slightly higher. The highest number of comments occur in the bin between 0 and 0.0625. The histogram is widely spread out indicating that comments have a wide variety of sentiment scores. Fig \ref{fig:commentlength} is a histogram that displays the distribution of comment length with a Kernel Density Estimate (KDE) curve. The x-axis represents the number of words in each comment. The y-axis represents the number of comments that belong to each bin. As the histogram is highly skewed to the left, the number of long comments is higher.

\begin{figure}
\includegraphics[width = 1\linewidth]{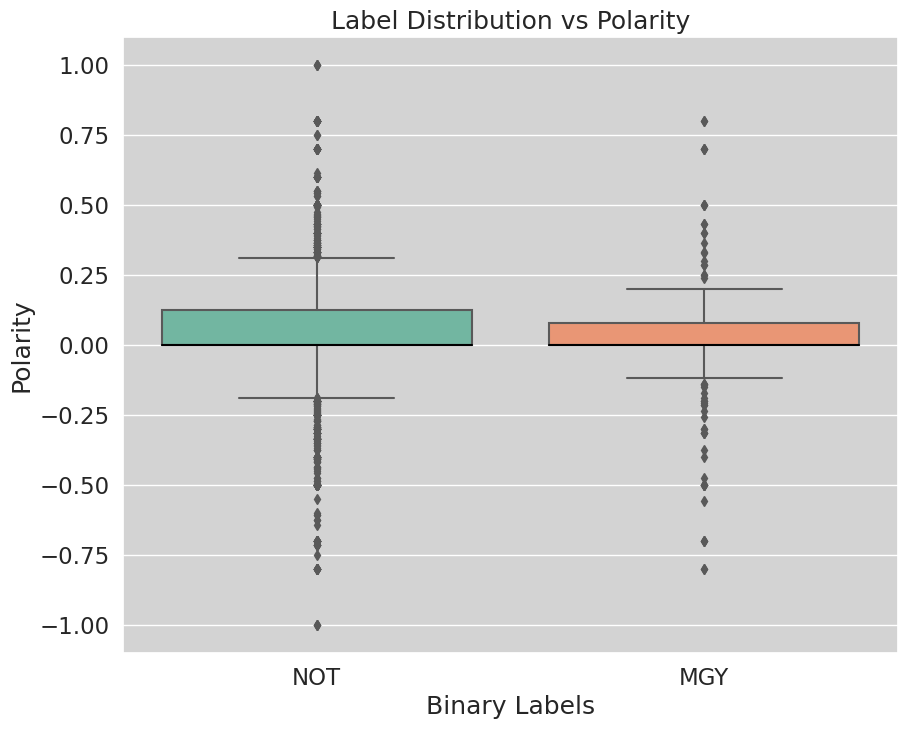}
    \caption{Boxplot of Binary Label Distribution and Polarity}
    \label{fig:boxplot}
\end{figure}

\begin{figure}
\includegraphics[width = 1\linewidth]{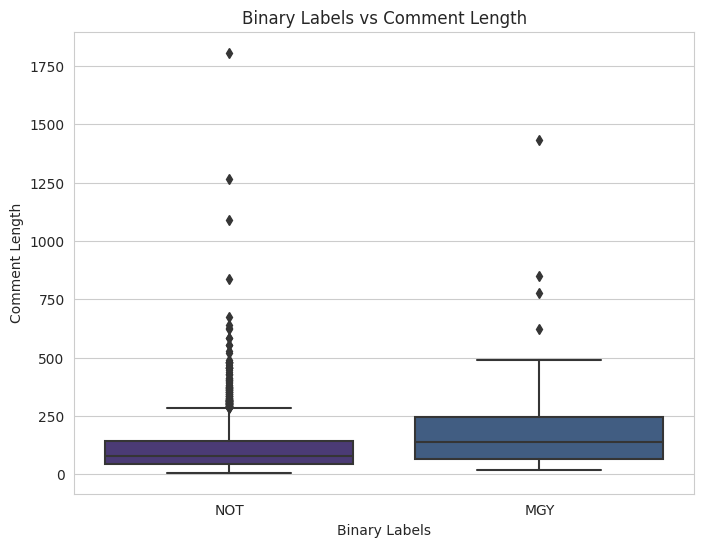}
    \caption{Boxplot of Binary Label Distribution and Comment Length}
    \label{fig:lenvslabel}
\end{figure}

Fig \ref{fig:boxplot} depicts a boxplot with a swarm plot overlay. The x-axis depicts the two binary classes, and the y-axis represents the sentiment scores. Each box presents the interquartile range (IQR).The small dots outside the box are individual comments. Fig \ref{fig:lenvslabel} displays a boxplot that depicts the binary labels (`NOT' and `MGY') on the x-axis and the comment length on the y-axis. The central tendency of the box depicting class `MGY' is higher than that of the box depicting class `NOT'. The box for class `MGY' is bigger than for `NOT', which indicates that the comment length is generally longer for class `NOT'. The plot also depicts an outlier in the class `NOT', with a comment of length of over 1750 characters.  

\subsection{Principal Component Analysis}
\begin{figure}
\centering\includegraphics[width = 0.8\linewidth]{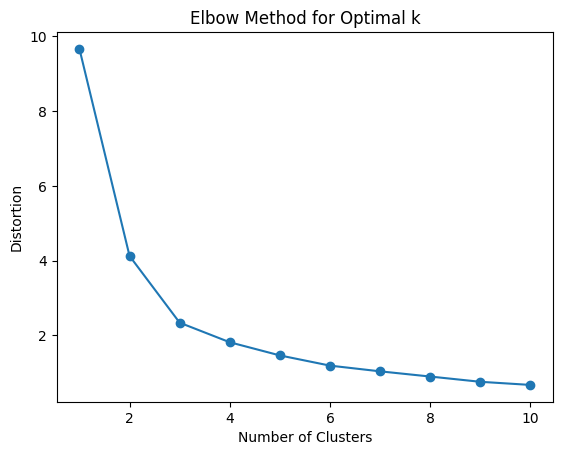}
    \caption{Elbow Method for PCA}
    \label{fig:elbowmethod}
\end{figure}

PCA has been performed on the dataset to reduce dimensionality of the data and find patterns. TF-IDF vectorizer with the maximum number of features = 3000 was used to obtain numerical features from the dataset. All the comments from the dataset have been selected for analysis. Fig \ref{fig:elbowmethod} displays a plot of the Elbow method, which shows that the optimal number of clusters was 3. 

\begin{figure}
\includegraphics[width = 1\linewidth]{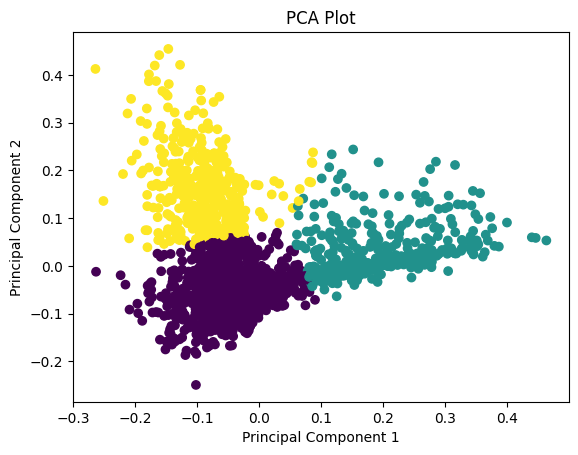}
    \caption{PCA Plot for All Comments}
    \label{fig:pca}
\end{figure}

The top keywords for cluster 1 are women, peopl, men, like, india, girl, feel, wear, this, indian. The top keywords for cluster 2 are hai, ko, ki, ke, bhi, se, ka, ye, aur, nahi. The top keywords for cluster 3 are rape, law, case, marit, consent, husband, wife, sex, women, men. Thus, cluster 2 consists mostly of mix-code tokens. An example of a comment in cluster 1 is: `Marriage doesn't mean only sex. It's emotional physical and spiritual'. An example of a comment in cluster 2 is: `Soch par fakrah hay !!  Savage!!' An example of a comment in cluster 3 is: `the fact there is not any law regarding such a big issue tells a lot about our country'.

\begin{table}[!ht]
    \centering
    \begin{tabular}{c|c|c|c}
\hline
         \textbf{Feature} & \textbf{Cluster 1} & \textbf{Cluster 2} & \textbf{Cluster 3} \\
         \hline
         Cluster Size &  466 &  1381 & 382\\ 
         Sentiment score & 0.0294 &  0.0137 & 0.0259 \\
         Average length & 16 & 26 & 30\\
         \hline
    \end{tabular}
    \caption{Cluster Characteristics using PCA}
    \label{tab:clustering}
\end{table}

Fig \ref{fig:pca} displays the PCA plot of the dataset which shows 3 distinct clusters. Table \ref{tab:clustering} displays scores for the 3 clusters obtained from the PCA plot. Cluster 2 has the highest number of comments and the lowest polarity score. This cluster contains mostly mix-code sentences and makes up the largest portion of the dataset. Cluster 3 has the highest positive sentiment score and the highest average length in words.  

\section{Discussion} \label{disc}
\begin{enumerate}
    \item Research Question 1: Can EDA techniques help uncover valuable insights from a dataset of Hinglish comments for the misogyny detection?
    In this paper, EDA techniques, such as comparison of samples across classes, PCA, keyword extraction, sentiment analysis, word cloud, and more, have been applied to the dataset. The analysis has provided valuable insights into the comment distribution across the dataset, polarity of comments, keywords, and more. Keyword extraction is an efficient strategy for hate speech classification through `hate' keywords \cite{saleh2023detection} \cite{rini2020systematic}.
    \item Research Question 2: What valuable insights and patterns are uncovered? Which EDA techniques provide valuable insights?
    The findings of the study suggest that misogynistic comments are generally longer than non-misogynistic comments. This aligns with previous studies that discuss sequence length and the corresponding class \cite{nozza2019unintended}. Most comments consist of less than 20 words, and most of the comments show slightly positive sentiment score using TextBlob.  
\end{enumerate}
\section{Conclusion} \label{conc}
This study was performed to conduct EDA on a code-mixed Hinglish dataset for misogyny detection. Several techniques such as word cloud, sentiment analysis, PCA, etc., were applied to the dataset. The findings have uncovered some interesting and useful correlations and patterns in the data, which can guide the modelling process. PCA obtained three clusters of comments, where code-mixed comments were clustered separately. % Correlation between the nine labels has shown that the there is positive correlation between Victim Blaming and Moral Policing. 
 Further analysis could reveal other important factors, and will be conducted alongside the training and evaluation of machine learning and deep learning models. In future studies, we will finish the fair annotation process with the Kappa coefficient and use machine learning algorithms on the dataset to investigate the implementation stage. 

%\section{References}

\bibliographystyle{unsrt}
\bibliography{references.bib}

\end{document}